\documentclass[twocolumn,subfig,verbose]{ceurart}

\usepackage{listings}
\usepackage{booktabs}
\usepackage{tabularx}
\usepackage{array}
\usepackage{booktabs}
\usepackage{microtype}
\begin{document}

\copyrightyear{2022}
\copyrightclause{Copyright for this paper by its authors.
  Use permitted under Creative Commons License Attribution 4.0
  International (CC BY 4.0).}

\conference{}

\title{From Tokens to Watt-hours: Analytical Energy Estimation for LLM Inference on Modern GPUs}


%
\author[1,2,3]{Tina Vartziotis}[%
orcid=0000-0002-0877-7063,
email=tina.vartziotis@twt-gmbh.de,
]
\cormark[1]
\fnmark[1]
\address[1]{National Technical University of Athens, Patission Complex 42, 10682 Athens, Greece}
\address[2]{Harvard University, 1350 Massachusetts Avenue, 02138 Cambridge, MA, USA}
\address[3]{TWT GmbH Science \& Innovation, Industriestraße 6, 70565 Stuttgart, DE}

\author[4]{Rodopi Kosteli}[%
orcid=0000-0001-7116-9338,
email=rodopi.kosteli@nikitec.gr,
]
\fnmark[1]
\address[4]{ NIKI Ltd Digital Engineering, 205 National Resistance Street, 45500 Ioannina, Greece}
\fnmark[1]
\address[5]{ University of Ioannina, Campus, 451 10 Ioannina, Greece}

\author[3,6]{Elli Danae Vartziotis}[%
orcid=0009-0006-1309-8598,
email=elli.vartziotis@twt-gmbh.de,
]

\address[6]{National and Kapodistrian University of Athens, Panepistimiou 30, 106 79 Athens, Greece}

\author[2]{George Dasoulas}[%
orcid=0000-0002-0562-5136,
email=gdasoulas@hsph.harvard.edu,
]
\author[3]{Michael Keckeisen}[
email=michael.keckeisen@twt-gmbh.de,
]
\author[5]{Konstantinos Skianis}[%
orcid=0000-0002-9421-8566,
email=kskianis@uoi.gr,
]

\author[1,7]{Sotirios Kotsopoulos}[%
orcid=0000-0002-9421-8566,
email=skots@mit.edu,
]
\address[7]{Massachusetts Institute of Technology, 77 Massachusetts Avenue, 02139 Cambridge, MA, USA}

\author[2]{Francesca Dominici}[%
orcid=0000-0002-9421-8566,
email=fdominic@hsph.harvard.edu,
]

\cortext[1]{Corresponding author.}
\fntext[1]{These authors contributed equally.}

\begin{abstract}
The operational energy consumption of large language model (LLM) inference is becoming an increasingly important component of the environmental footprint of deployed AI systems. However, direct measurement of inference energy often requires hardware telemetry, power instrumentation, or infrastructure-specific monitoring, limiting its applicability in comparative studies, early-stage system design, and sustainability reporting. This report presents an analytically structured, empirically calibrated, GPU-level methodology for estimating LLM inference energy on NVIDIA H100-class accelerators without direct runtime measurement. The proposed estimator combines parameter-scaled transformer FLOP accounting, calibrated memory-traffic factors, and hardware-specific energy coefficients for FP16/BF16 tensor-core computation and high-bandwidth-memory movement. It explicitly separates prompt prefill from autoregressive decoding, enabling energy estimates for input tokens, output tokens, and complete inference requests. The methodology further decomposes total energy into compute, parameter-access, key--value-cache write, and attention-read components, allowing the scaling behavior with model size, context length, and generated-token count to be analyzed. The resulting estimates are not intended to replace physical power measurements; rather, they provide transparent, reproducible, and assumption-explicit approximations suitable for model comparison, green-coding analysis, and design-time evaluation of LLM inference workloads.
\end{abstract}


\maketitle

\section{Introduction}

As large language models (LLMs) scale to serve millions of users across cloud, edge, and on-premise deployments, their cumulative inference energy has emerged as a first-order sustainability concern, driven by the substantial computational and environmental costs of large-scale AI inference. Prior work in Green AI has shown that advances in model capability are often accompanied by increases in energy consumption and carbon emissions, motivating energy-aware approaches to machine learning research and deployment~\cite{strubell2019energy,schwartz2020green}.

While early work focused primarily on training costs \cite{patterson2021carbon}, broader Green AI research highlighted the importance of computational efficiency, energy-aware reporting, and environmental impact across the ML lifecycle \cite{schwartz2020green, henderson2020systematic, lacoste2019quantifying, lannelongue2021green}, including recent efforts to extend energy-aware optimization to model selection \cite{betello2025one}. Attention has since shifted toward inference, which is becoming an increasingly important operational component of deployed LLM systems. Unlike training, which is performed occasionally, inference workloads run continuously and scale with user demand through autoregressive token generation, accumulating substantial energy costs over a model's deployment lifetime \cite{lim2024carbon}. Their energy footprint depends strongly on generated-token count, sequence length, model size, hardware configuration, and batching behavior \cite{fernandez2025energy, fu2024llmco2, Vartziotis_LLMasaService}. 

This makes inference energy a practical concern not only for large cloud providers, but also for applied AI teams, research groups, and small-to-medium organizations deploying LLMs on local or on-premise GPU infrastructure. For such deployments, accelerator-side energy does not capture the full service-level cost, since system and facility overheads, including cooling, networking, and power delivery, also contribute \cite{uptime2024survey, mlperfpower2024}. Nevertheless, GPU-side energy is often a major controllable component of inference cost in GPU-dense systems \cite{nvidia_h100_datasheet}, making GPU-level estimation useful for comparing model and workload choices, assessing green-coding interventions \cite{Vartziotis_GreenCode}, and making deployment-time decisions when full datacenter-level instrumentation is not available.

Existing approaches for estimating or reporting AI-related energy consumption often rely on hardware telemetry, external power measurement, infrastructure-specific monitoring, or coarse-grained carbon-accounting tools \cite{lannelongue2021green, lacoste2019quantifying}. Measurement-based studies provide valuable empirical evidence  across hardware configurations and task types\cite{samsi_words_2023, luccioni_power_2024}, but their results are tied to specific hardware, serving systems, and runtime configurations. Carbon-aware inference work such as SPROUT further demonstrates that autoregressive generation can itself be optimized for sustainability \cite{li2024sprout}. Conversely, transformer scaling-law studies and GPU microarchitectural benchmarks provide the basis for analytical FLOP and hardware-energy modeling \cite{kaplan2020scaling, hoffmann2022training, antepara2025benchmark}, but do not by themselves provide request-level or token-level inference energy estimates. This leaves a gap for transparent, reproducible estimators that combine transformer compute modeling, memory-traffic attribution, and hardware-specific energy coefficients to estimate request-level and normalized token-level GPU inference energy without requiring runtime instrumentation.

We address this gap by presenting a semi-analytical methodology for estimating accelerator-side GPU compute and memory-movement energy for LLM inference. The estimator combines parameter-scaled transformer FLOP modeling, calibrated HBM memory-traffic estimation, and hardware-specific energy coefficients. It separates prompt prefill from autoregressive decode, reports normalized input-token, output-token, and request-level energy estimates, and decomposes total energy into compute, parameter-access, KV-cache-write, and attention-read components. Beyond comparative estimation, the framework also highlights actionable mechanisms for reducing GPU inference energy, including smaller models, shorter generations, KV-cache quantization, prompt compression, and improved batching efficiency.

\section{Methodology}
\label{sec:analytical_methodology}

This section defines the analytically structured, empirically calibrated
methodology used to estimate GPU-level energy consumption for LLM inference. The estimator is designed for settings in
which direct runtime power measurement is unavailable, impractical, or not
comparable across systems. It therefore provides assumption-explicit energy
estimates rather than replacing hardware-level measurement.
The scope of the estimator is accelerator-side operational energy. We model only
the energy associated with GPU-side computation and GPU memory movement during
inference. System-level and datacenter-level contributions, including CPU
execution, host memory, networking, storage, power-supply losses, cooling, and
power usage effectiveness (PUE), are outside the scope of this work and are not
included in the reported estimates. Similarly, this paper does not estimate
carbon emissions; all reported quantities are expressed as GPU-level energy.

Each inference request is
separated into two phases: prompt prefill and autoregressive decode.  During prefill, the model processes the input prompt and constructs the initial
key--value (KV) cache. During decode, the model generates output tokens
sequentially, with each step attending to the accumulated context through the KV
cache. This phase separation is required because input-token and output-token
costs have different compute and memory-access patterns.

\subsection{Request-Level Energy Decomposition}

Let $T_{\mathrm{in}}$ denote the number of input tokens, i.e., the tokenized prompt provided to the model, and let $T_{\mathrm{out}}$ denote the number of generated output tokens. 
The total GPU energy of one request, denoted $E_{\mathrm{GPU}}$, is
decomposed into prefill and decode energy:
\begin{equation}
E_{\mathrm{GPU}}
=
E_{\mathrm{pre}}
+
E_{\mathrm{dec}}.
\label{eq:phase_decomposition}
\end{equation}
where $E_{\mathrm{pre}}$ is the energy consumed during the prefill phase and
$E_{\mathrm{dec}}$ is the energy consumed during the autoregressive decode phase.
The reported input-token and output-token energy values are defined as:
\begin{equation}
E_{\mathrm{in/token}}
=
\frac{E_{\mathrm{pre}}}{T_{\mathrm{in}}},
\qquad
E_{\mathrm{out/token}}
=
\frac{E_{\mathrm{dec}}}{T_{\mathrm{out}}}.
\label{eq:token_energy}
\end{equation}
where $E_{\mathrm{in/token}}$ and $E_{\mathrm{out/token}}$ denote the average
energy per input token and per generated output token, respectively. These quantities are not intrinsic constants of a model. They depend on prompt length, generated-token count, inference precision, batching, cache reuse, hardware characteristics, and serving implementation. At the GPU level, request energy is modeled as the sum of tensor-core compute
energy and high-bandwidth-memory (HBM) movement energy:
\begin{equation}
E_{\mathrm{GPU}}
=
E_{\mathrm{compute}}
+
E_{\mathrm{memory}}.
\label{eq:gpu_decomposition}
\end{equation}
where $E_{\mathrm{compute}}$ denotes energy consumed by tensor-core
computation, and $E_{\mathrm{memory}}$ denotes energy consumed by memory movement.
This decomposition follows common GPU energy modeling approaches,
which treat computation and data movement as the dominant
contributors to energy consumption~\cite{antepara2025benchmark}.
The compute term is proportional to the number of tensor-core floating-point
operations:
\begin{equation}
E_{\mathrm{compute}}
=
\alpha_{\mathrm{TC}} C_{\mathrm{TC}},
\label{eq:compute_energy}
\end{equation}
where $C_{\mathrm{TC}}$ denotes tensor-core FLOPs and $\alpha_{\mathrm{TC}}$ is
the hardware-specific energy per tensor-core FLOP.
The memory term is proportional to the number of bits transferred through HBM:
\begin{equation}
E_{\mathrm{memory}}
=
e_{\mathrm{HBM}} Q_{\mathrm{HBM}},
\label{eq:memory_energy}
\end{equation}
where $Q_{\mathrm{HBM}}$ denotes HBM traffic in bits and $e_{\mathrm{HBM}}$ is
the energy per transferred HBM bit.
Combining these terms gives:
\begin{equation}
E_{\mathrm{GPU}}
=
\alpha_{\mathrm{TC}} C_{\mathrm{TC}}
+
e_{\mathrm{HBM}} Q_{\mathrm{HBM}}.
\label{eq:main_energy_model}
\end{equation}

This formulation separates workload-dependent quantities
$(C_{\mathrm{TC}}, Q_{\mathrm{HBM}})$ from hardware-specific coefficients $(\alpha_{\mathrm{TC}}, e_{\mathrm{HBM}})$~\cite{antepara2025benchmark}.

\subsection{Compute Model}
\label{subsec:compute_model}

For dense decoder-only transformers, we approximate the dominant tensor-core
FLOPs using the standard parameter-scaled transformer estimate:
\begin{align}
C_{\mathrm{pre,dense}} &= K N T_{\mathrm{in}},
&
C_{\mathrm{dec,dense}} &= K N T_{\mathrm{out}},
\label{eq:dense_compute_main}
\end{align}
where $N$ is the number of model parameters and $K=6$ is the FLOP coefficient
per parameter per token. For long contexts, architecture-aware attention corrections are added through
\begin{equation}
C_{\mathrm{pre}} =
C_{\mathrm{pre,dense}} + C_{\mathrm{pre,attn}},
\qquad
C_{\mathrm{dec}} =
C_{\mathrm{dec,dense}} + C_{\mathrm{dec,attn}},
\label{eq:phase_compute_main}
\end{equation}
so that
\begin{equation}
C_{\mathrm{TC}} =
C_{\mathrm{pre}} + C_{\mathrm{dec}}.
\label{eq:total_compute_main}
\end{equation}

The full attention correction terms are provided in Supplementary
Section~\ref{supp_sec:compute_derivation}.

\subsection{Memory-Movement Model}
\label{subsec:memory_model}

In addition to tensor-core computation, LLM inference requires substantial data movement through the GPU memory hierarchy. We approximate the dominant off-chip memory traffic as high-bandwidth-memory (HBM) traffic and decompose it into  parameter-access traffic, KV-cache write traffic, and attention-related KV-cache read traffic:
\begin{equation}
\mathrm{Bits}_{\mathrm{total}}
=
\mathrm{Bits}_{\mathrm{params}}
+
\mathrm{Bits}_{\mathrm{KV}}
+
\mathrm{Bits}_{\mathrm{attn}}'.
\label{eq:bits_total}
\end{equation}
This decomposition captures the primary sources of memory traffic in autoregressive transformer inference, including parameter
access, KV-cache storage, and attention-related reads, as
observed in modern LLM serving systems~\cite{kwon2023efficient}.
The corresponding memory energy is modeled as
\begin{equation}
E_{\mathrm{memory}}
=
\mathrm{Bits}_{\mathrm{total}}
\cdot
e_{\mathrm{HBM}}
\cdot
\eta(N),
\label{eq:memory_energy_calibrated}
\end{equation}
where $e_{\mathrm{HBM}}$ is the energy per bit transferred from HBM and $\eta(N)$ is a calibrated memory-inefficiency factor. The individual memory-traffic terms are defined in Section~\ref{subsec:fixed_model_parameters}, with full derivations provided in Supplementary Section~\ref{supp_sec:memory_derivation}.

\subsection{Total Request and Token-Level Energy}
\label{subsec:total_request_energy}

The total GPU energy for a request, denoted
$E_{\mathrm{request}}$, is obtained by combining the tensor-core
compute term and the calibrated HBM memory-movement term:
\begin{equation}
E_{\mathrm{request}}
=
\alpha_{\mathrm{TC}}
\left(
C_{\mathrm{pre}} + C_{\mathrm{dec}}
\right)
+
\mathrm{Bits}_{\mathrm{total}}
\cdot
e_{\mathrm{HBM}}
\cdot
\eta(N).
\label{eq:request_energy_calibrated}
\end{equation}
where $C_{\mathrm{pre}}$ and $C_{\mathrm{dec}}$ denote the total
prefill and decode tensor-core FLOPs, obtained from the dense terms in
Section~\ref{subsec:compute_model} and, when applicable, the attention
correction terms in Supplementary Section~\ref{supp_sec:compute_derivation}. The term $\mathrm{Bits}_{\mathrm{total}}$ denotes total HBM traffic
(Equation~\ref{eq:bits_total}). The factor $\eta(N)$ is a
dimensionless inefficiency multiplier applied to the memory term,
representing non-ideal HBM behavior. The average energy per processed token is:
\begin{equation}
E_{\mathrm{avg/token}}
=
\frac{
E_{\mathrm{request}}
}{
T_{\mathrm{in}} + T_{\mathrm{out}}
}.
\label{eq:avg_token_energy}
\end{equation}
where $T_{\mathrm{in}}$ and $T_{\mathrm{out}}$ are the numbers of
input and generated output tokens, respectively. For phase-level reporting, we compute:
\begin{equation}
E_{\mathrm{pre}}
=
E_{\mathrm{compute,pre}}
+
E_{\mathrm{memory,pre}},
\qquad
E_{\mathrm{dec}}
=
E_{\mathrm{compute,dec}}
+
E_{\mathrm{memory,dec}}.
\label{eq:phase_components}
\end{equation}
where $E_{\mathrm{compute,pre}}$ and $E_{\mathrm{compute,dec}}$
denote the tensor-core compute energy consumed during the prefill
and decode phases, respectively, and $E_{\mathrm{memory,pre}}$ and
$E_{\mathrm{memory,dec}}$ denote the corresponding memory-movement
energy. The input-token and output-token metrics are then obtained using Equation~\ref{eq:token_energy}.

\subsection{Hardware Coefficients}

The hardware coefficients convert tensor-core FLOPs and HBM traffic into energy.
In the main evaluation, we instantiate the estimator for H100-class FP16/BF16
inference using the fixed coefficients reported in
Table~\ref{tab:constants}:
\(\alpha_{\mathrm{TC}} = 0.52\) pJ/FLOP and
\(e_{\mathrm{HBM}} = 11.68\) pJ/bit. These values are based on microarchitectural accelerator-energy measurements  of GPU tensor-core computation and HBM access reported by Antepara et al.~\cite{antepara2025benchmark}.
These coefficients represent microarchitectural accelerator energy rather than
wall-plug or datacenter-level energy. They are
GPU-level compute and HBM-transfer coefficients. Additional coefficients for other accelerator classes, such as A100, are
provided in the Supplementary Table~\ref{supp_tab:hardware_coefficients}.

\subsection{Simplified Parameter-Only Estimator}

For model inventories where detailed architecture information is unavailable, we
also define a simplified estimator. In this case, output-token energy is
approximated as:
\begin{equation}
\widehat{E}_{\mathrm{out/token}}
=
\alpha_{\mathrm{eff}} K N,
\label{eq:simplified_output_token}
\end{equation}
where $\alpha_{\mathrm{eff}}$ is an effective energy-per-FLOP coefficient that absorbs compute, memory, and utilization effects.
This parameter-scaled approximation is consistent with prior
analyses showing that transformer compute scales linearly with model size~\cite{kaplan2020scaling,hoffmann2022training}.
Input-token energy is modeled as a prompt-length-dependent multiple of
output-token energy:
\begin{equation}
\widehat{E}_{\mathrm{in/token}}
=
M(T_{\mathrm{in}})
\alpha_{\mathrm{eff}} K N.
\label{eq:simplified_input_token}
\end{equation}

The request-level estimate is:
\begin{equation}
\widehat{E}_{\mathrm{request}}
=
T_{\mathrm{in}}
\widehat{E}_{\mathrm{in/token}}
+
T_{\mathrm{out}}
\widehat{E}_{\mathrm{out/token}}.
\label{eq:simplified_request}
\end{equation}

where $M(T_{\mathrm{in}})$ is a prompt-length-dependent prefill
multiplier. Its functional form is defined in
Section~\ref{subsec:fixed_model_parameters}.

The simplified estimator is useful when only parameter counts and token counts
are known. The architecture-aware estimator in equation~\ref{eq:request_energy_calibrated}
should be preferred whenever layer count, hidden dimension, KV-cache dimension,
precision, and workload assumptions are available.

\section{Evaluation Setup and Assumptions}
\label{sec:evaluation_setup}

This section specifies the hardware configuration, model set, workload
assumptions, and fixed estimator parameters used in the analytical evaluation.
The purpose is to make the reported energy estimates reproducible and to
separate methodological assumptions from the numerical results.

\subsection{Target Hardware and Precision}

The evaluation is instantiated for NVIDIA H100-class accelerators, corresponding
to the inference hardware considered in this work. Unless otherwise stated, all
models are assumed to run using FP16 or BF16 tensor-core inference. Model weights
and KV-cache entries are therefore assumed to use 16-bit precision:
\begin{equation}
b_w = b_{\mathrm{kv}} = 16.
\end{equation}
For the H100-class configuration, we use the accelerator-level energy
coefficients summarized in Table~\ref{tab:constants}. Here, $\alpha_{\mathrm{TC}}$ denotes the energy per FP16/BF16 tensor-core FLOP
and $e_{\mathrm{HBM}}$ denotes the energy per bit transferred through
high-bandwidth memory. These coefficients represent GPU-level
microarchitectural energy and do not include CPU, cooling, networking, or
datacenter-level overheads.

\subsection{Model Set}

We evaluate a representative set of transformer-based LLMs that cover small ($< 3B$), medium ($3B-30B$), and large ($> 30B$) parameter regimes as summarized in Supplementary Table~\ref{supp_tab:model_presets}. The model set includes embedding models,
general-purpose decoder-only LLMs, code models, reasoning models, and
vision-language models. For each model, the estimator requires at minimum the
parameter count $N$ and the input/output token counts. When available, additional
architecture-specific quantities such as the number of layers $n_{\ell}$, hidden
dimension $d_{\mathrm{model}}$, Kv
and effective KV-cache dimension
$d_{\mathrm{kv}}$ are used by the architecture-aware estimator. If full architecture metadata is unavailable, the simplified parameter-only
estimator is used. In this case, the model is represented by its parameter count
and token workload only. This enables consistent comparison across heterogeneous
model inventories while preserving explicit assumptions.

\subsection{Workload Scenarios}

The analytical evaluation considers request-level inference workloads defined by
the number of input tokens $T_{\mathrm{in}}$ and generated output tokens
$T_{\mathrm{out}}$. The default workload used for model-size comparisons is:
\begin{equation}
T_{\mathrm{in}} = 500,
\qquad
T_{\mathrm{out}} = 500.
\end{equation}
Additional experiments vary $T_{\mathrm{out}}$ while holding model size and
input length fixed in order to study how inference energy scales with generated
sequence length. This isolates the effect of autoregressive decoding and
attention-related KV-cache reads. Unless otherwise stated, the evaluation assumes
single-request execution and does not explicitly model continuous batching.
Effects of cache reuse, locality, and batching on parameter-access traffic are
represented through the parameter-access factor $\gamma$.

\subsection{Fixed Model Parameters}
\label{subsec:fixed_model_parameters}

We adopt standard constants for transformer inference, including the dense transformer FLOP coefficient $K$, along with hardware-related energy parameters. All fixed modeling constants and hardware coefficients are summarised in Table~\ref{tab:constants}.

\begin{table*}[t]
\centering
\caption{Fixed constants and hardware coefficients used in the evaluation.}
\label{tab:constants}

\begin{tabularx}{\textwidth}{
    >{\raggedright\arraybackslash}p{0.18\textwidth}
    >{\raggedright\arraybackslash}p{0.20\textwidth}
    >{\raggedright\arraybackslash}X
}
\toprule
\textbf{Parameter} & \textbf{Value} & \textbf{Description} \\
\midrule
$K$ & 6 &
FLOPs per parameter per token (transformer constant) \\

$\alpha_{\mathrm{TC}}$ & 0.52 pJ/FLOP &
H100 tensor-core energy coefficient \\

$e_{\mathrm{HBM}}$ & 11.68 pJ/bit &
HBM energy per bit transferred \\

$b_w$ & 16 &
Bits per model weight (FP16/BF16) \\

$b_{\mathrm{kv}}$ & 16 &
Bits per KV-cache element \\
\bottomrule
\end{tabularx}
\end{table*}

The individual memory-traffic terms in Equation~\ref{eq:bits_total} are
specified by the parameter-access, KV-cache-write, and attention-read terms
below:
\begin{align}
\mathrm{Bits}_{\mathrm{params}} &= b_w N \gamma(N),
\hspace{1.5em}
\gamma(N) = \gamma_0\left(\frac{N}{N_0}\right)^\beta,
\label{eq:gamma_scaling_main}
\\
\mathrm{Bits}_{\mathrm{KV}} &=
2 b_{\mathrm{kv}} d_{\mathrm{model}} n_{\ell} T_{\mathrm{out}},
\label{eq:bits_kv_main}
\\
\mathrm{Bits}_{\mathrm{attn}}' &=
\left[
2 b_{\mathrm{kv}} d_{\mathrm{model}} n_{\ell}
\left(
T_{\mathrm{out}}T_{\mathrm{in}}
+
\frac{T_{\mathrm{out}}(T_{\mathrm{out}}-1)}{2}
\right)
\right]
s_{\mathrm{attn}}(N).
\label{eq:bits_attn_scaled_main}
\end{align}
Here, $\gamma(N)$ captures effective parameter reuse, while
$s_{\mathrm{attn}}(N)$ captures non-ideal KV-cache read overhead; their calibrated
forms are given in Table~\ref{tab:model_parameters}. Full derivations and interpretation are provided in Supplementary
Section~\ref{supp_sec:memory_derivation}.

For the simplified parameter-only estimator, the prefill multiplier $M(T_{\mathrm{in}})$ is selected according to the input length bucket:

\begin{equation}
M(T_{\mathrm{in}}) =
\begin{cases}
1.2, & T_{\mathrm{in}} \le 2048, \\
1.8, & 2048 < T_{\mathrm{in}} \le 5120, \\
3.0, & 5120 < T_{\mathrm{in}} \le 10240, \\
4.0, & T_{\mathrm{in}} > 10240.
\end{cases}
\end{equation}

This multiplier is used only in the simplified parameter-only estimator.

\subsection{Calibration Procedure}
\label{subsec:calibration_procedure}
We estimate the model parameters using a data-driven calibration
procedure based on reported energy measurements for LLM inference~\cite{caravaca2025prompts}.
For each model size, we first decompose the total energy into
compute and memory components, using the analytical expressions
derived in the previous sections. The remaining memory contribution
is then further decomposed into parameter, KV-cache, and attention
terms. Using this decomposition, we obtain approximate estimates of the
effective parameter-access factor $\gamma(N)$ for each model:
\begin{equation}
\gamma(N) =
\frac{\mathrm{Bits}_{\mathrm{params}}}{b_w N},
\end{equation}
where $\mathrm{Bits}_{\mathrm{params}}$ is inferred from measured
energy after subtracting compute and other memory contributions. The parameters $\gamma_0$ and $\beta$ are then determined by fitting
the power-law model to these inferred values, while also ensuring
consistency with the overall energy estimates. The parameters of $s_{\mathrm{attn}}(N)$ (Equation~\ref{eq:bits_attn_scaled_main})
and $\eta(N)$ (Equation~\ref{eq:memory_energy_calibrated}) are calibrated by minimizing the deviation between model predictions and measurement-based reported energy values across the evaluated models. In practice, we perform
a low-dimensional parameter search over the coefficients of the scaling
functions, selecting values that minimize the relative error while
preserving the expected scaling trends with model size.

This procedure is a constrained least-squares fitting over a small number of parameters, where the objective is to match both
the magnitude and growth behavior of measured energy consumption rather
than exactly fitting individual data points. Due to the small number of calibration parameters and limited data points,
the calibration is intentionally kept simple to avoid overfitting and to
retain generalizability across models and workloads. After calibration, these factors are kept fixed across the scaling and decomposition experiments. 

This calibration makes the estimator an analytically structured, 
empirically calibrated model: transformer FLOP counts and memory-traffic terms
provide the analytical structure, while the calibrated factors represent
non-ideal reuse, attention-access overhead, and global HBM inefficiency. The
factors should therefore be recalibrated when applying the methodology to a
different GPU generation, inference engine, batching regime, or serving
configuration.

\begin{table*}[t]
\centering
\caption{Calibrated model parameters used in the analytical estimator.}
\label{tab:model_parameters}

\begin{tabularx}{\textwidth}{
    >{\raggedright\arraybackslash}p{0.16\textwidth}
    >{\raggedright\arraybackslash}p{0.30\textwidth}
    >{\raggedright\arraybackslash}X
}
\toprule
\textbf{Parameter} & \textbf{Value} & \textbf{Description} \\
\midrule

$\gamma_0$ &
0.10 &
Baseline parameter-access factor at reference model size $N_0$ \\

$N_0$ &
24B &
Reference model size for scaling relationships \\

$\beta$ &
0.8 &
Exponent controlling degradation of parameter reuse with model size \\

$s_{\mathrm{attn}}(N)$ &
$\displaystyle 1 + 1.5\left(\frac{N}{N_0}\right)^{0.9}$ &
Attention scaling factor capturing KV-cache read overhead \\

$\eta(N)$ &
$\displaystyle 1 + 0.8\left(\frac{N}{N_0}\right)^{0.8}$ &
Global memory-inefficiency factor for HBM traffic \\

\bottomrule
\end{tabularx}
\end{table*}

\subsection{Reported Metrics}
\label{subsec:reported_metrics}

For each model--workload pair, the estimator reports prefill energy
$E_{\mathrm{pre}}$, decode energy $E_{\mathrm{dec}}$, total request energy
$E_{\mathrm{request}}$, input-token energy, output-token energy, and average
energy per processed token. The full description of these metrics is provided in Supplementary
Section~\ref{supp_sec:reported_metrics}.

\subsection{Interpretation of Estimates}

All reported values are analytical GPU-level estimates and are scenario-dependent approximations rather than direct measurements
or intrinsic properties of the models. Differences between measured and
estimated energy may arise from batching, tensor-parallel communication,
framework overheads, kernel fusion, quantization, cache behavior, and runtime GPU
utilization. These effects are discussed further in the limitations section.

\section{Results and Evaluation}
\label{sec:results}

This section presents the analytical energy estimates obtained using the setup
defined in Section~\ref{sec:evaluation_setup}.  We evaluate the estimator along
four dimensions: scaling with generated sequence length, scaling with model
size, decomposition of energy into compute and memory components, and comparison
against measurement-based results from prior work. We also report token-level
energy estimates for the model inventory considered in this study. Unless otherwise stated, all results assume FP16/BF16 inference on H100-class accelerators, a dense transformer FLOP constant $K=6$, the calibrated
parameter-access factor $\gamma(N)$, the attention-access factor
$s_{\mathrm{attn}}(N)$, the memory-inefficiency factor $\eta(N)$, and the
hardware coefficients $\alpha_{\mathrm{TC}} = 0.52$ pJ/FLOP and
$e_{\mathrm{HBM}} = 11.68$ pJ/bit. The set of models considered in this study and their key architectural characteristics are summarized in Table~S\ref{supp_tab:model_presets}, which can be found in the Supplementary material.

\subsection{Scaling with Generated Output Length}

We first evaluate how inference energy scales with the number of generated
tokens. For this experiment, we fix the model size to 32B parameters and use an
input length of $T_{\mathrm{in}} = 100$ tokens. We then vary the number of
generated output tokens $T_{\mathrm{out}}$.

\begin{figure}[h]
    \centering
    \includegraphics[width=1\linewidth]{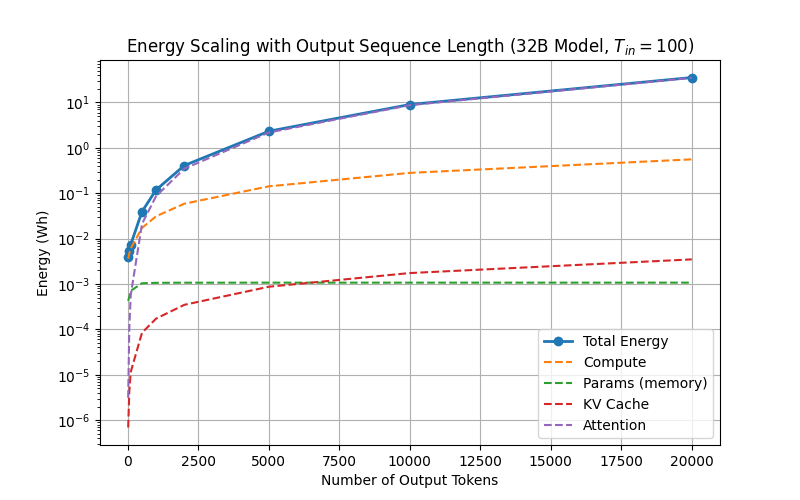}
    \caption{
    Energy breakdown as a function of generated output length
    $T_{\mathrm{out}}$ for a 32B-parameter model with
    $T_{\mathrm{in}}=100$ input tokens. Compute and KV-cache write costs scale
    almost linearly with $T_{\mathrm{out}}$, while attention-related
    KV-cache reads exhibit quadratic growth and become
increasingly significant at larger output lengths.
    }
    \label{fig:scaling_output_tokens}
\end{figure}

Figure~\ref{fig:scaling_output_tokens} shows that compute energy grows
approximately linearly with $T_{\mathrm{out}}$, consistent with the
autoregressive decoding process in which each generated token requires one
forward pass through the model. KV-cache write traffic also grows linearly,
because each generated token contributes one new key and value entry per layer. In contrast, scaled attention-related memory traffic grows super-linearly. This
is due both to the quadratic growth of KV-cache reads with output length and to
the attention-access factor $s_{\mathrm{attn}}(N)$ used in the calibrated memory
model. During decode, each newly generated token attends over the accumulated
context, so the total number of KV-cache reads increases as:
\begin{equation}
T_{\mathrm{out}}T_{\mathrm{in}}
+
\frac{T_{\mathrm{out}}(T_{\mathrm{out}}-1)}{2}.
\end{equation}

\subsection{Scaling with Model Size}

We next evaluate the dependence of request-level energy on model size. The
workload is fixed to:
\begin{equation}
T_{\mathrm{in}} = 500,
\qquad
T_{\mathrm{out}} = 500.
\end{equation}

This isolates the effect of model parameter count while holding the token
workload constant.

\begin{figure}[ht]
    \centering
    \includegraphics[width=1\linewidth]{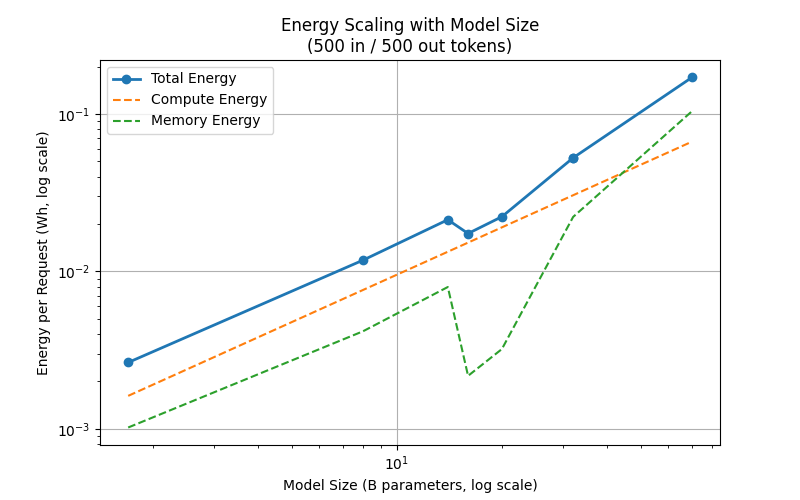}
    \caption{
    Energy per request as a function of model size for a fixed workload of
    500 input tokens and 500 output tokens. Total energy, compute energy, and
    memory energy are shown on a log--log scale.
    }
    \label{fig:model_scaling}
\end{figure}

Figure~\ref{fig:model_scaling} shows that total request energy increases
approximately linearly with model size on a log--log scale. This behavior is
expected from the parameter-scaled compute model:
\begin{equation}
C_{\mathrm{dec,dense}}
=
KNT_{\mathrm{out}},
\qquad
C_{\mathrm{pre,dense}}
=
KNT_{\mathrm{in}}.
\end{equation}

For fixed input and output lengths, the dominant dense matrix operations scale
linearly with the number of parameters $N$. Consequently, the compute component
follows:
\begin{equation}
E_{\mathrm{compute}} \propto N.
\end{equation}

The memory component increases with model size through the calibrated factor $\gamma(N)$, the attention-access factor
$s_{\mathrm{attn}}(N)$, and the global memory-inefficiency factor $\eta(N)$.
Under the fixed 500-token/500-token workload, compute remains the dominant
component for most evaluated model sizes, although memory energy increases
nonlinearly with the calibrated memory factors. While the overall scaling trend is approximately linear in model size,
the memory-energy component exhibits non-monotonic behavior for certain
models. This is due to differences in architectural configurations,
rather than parameter count alone. In particular, key contributors to
memory traffic, such as the hidden dimension $d_{\mathrm{model}}$ and
the number of layers $n_{\ell}$, do not scale uniformly with $N$
across different models. As a result, some medium-sized models may exhibit lower memory traffic
than nearby models with slightly different parameter counts,
leading to localized deviations (e.g., a reduction in the memory-energy
curve). This highlights that memory costs are sensitive to architectural
design choices, not solely parameter count.

\subsection{Energy Decomposition}

\begin{figure}[th]
    \centering
    \includegraphics[width=1\linewidth]{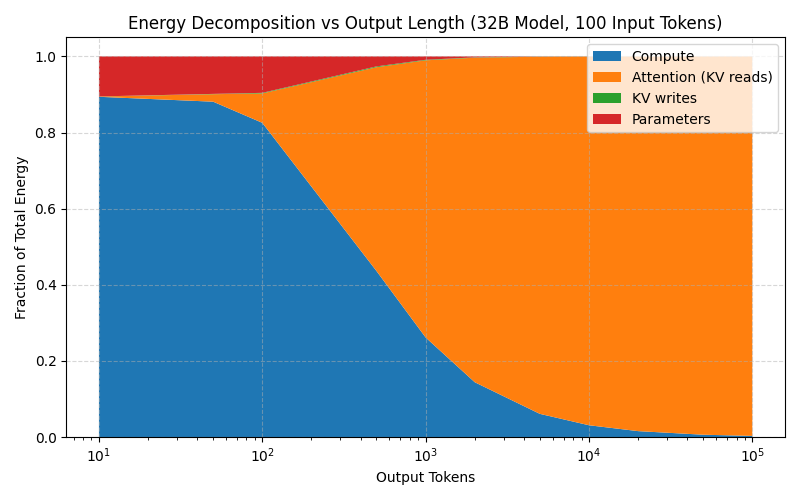}
    \caption{
    Fractional energy contribution as a function of generated output length
    for a 32B-parameter model. Compute dominates at short and moderate output
    lengths, while attention-related memory traffic increases with sequence
    length due to quadratic KV-cache read scaling.
    }
    \label{fig:energy_decomposition}
\end{figure}

To understand which terms dominate request energy, we decompose total energy into compute, parameter movement, KV-cache writes, and attention-related KV-cache reads. The decomposition is evaluated for a 32B-parameter model while varying $T_{\mathrm{out}}$. Figure~\ref{fig:energy_decomposition} shows the absolute energy
decomposition as a function of generated output length for a 32B-parameter
model. The stacked representation highlights the additive structure of the
energy model, where total energy is the sum of compute and memory components.

Compute energy increases approximately linearly with $T_{\mathrm{out}}$, reflecting the constant per-token cost of autoregressive decoding.
In contrast, attention-related memory energy exhibits super-linear growth, due to the quadratic scaling of KV-cache reads with sequence length. At small output lengths, total energy is dominated by compute. As $T_{\mathrm{out}}$ increases, the attention-related component grows rapidly and becomes a substantial contributor to the overall energy. This results in an upward curvature of the total energy trend, indicating the increasing impact of memory movement at longer sequence lengths.

The contributions from parameter access and KV-cache writes are negligible relative to compute and attention-related memory traffic across the evaluated range. In particular, the KV-cache write component, although included in the model, is not visually
distinguishable in the stacked representation. This is because KV writes scale linearly with $T_{\mathrm{out}}$, while
attention-related KV-cache reads scale quadratically and dominate memory traffic at larger sequence lengths.

\subsection{Token-Level Energy Estimates for the Model Inventory}

Table~\ref{tab:model_inventory_token_energy} reports simplified token-level
estimates for the model inventory using the parameter-only compute-dominated
estimator:
\begin{equation}
\widehat{E}_{\mathrm{out/token}}
=
\alpha_{\mathrm{TC}} K N,
\qquad
\widehat{E}_{\mathrm{in/token}}
\approx
1.2\,\widehat{E}_{\mathrm{out/token}},
\end{equation}
where the input-token multiplier corresponds to the short-prompt setting. These values are simplified, compute-dominated estimates; further interpretation is provided in Supplementary Section~\ref{supp_sec:reported_metrics}.

\begin{table*}[t]
\centering
\caption{Estimated energy consumption per token and per request across evaluated models.}
\label{tab:model_inventory_token_energy}
\small
\setlength{\tabcolsep}{8pt}
\renewcommand{\arraystretch}{1.08}

\begin{tabular}{lrrrr}
\toprule
\textbf{Model} &
\textbf{Params (B)} &
\textbf{$E_{\mathrm{out/token}}$} &
\textbf{$E_{\mathrm{in/token}}$} &
\textbf{$E_{\mathrm{request}}$} \\
&
&
\textbf{(mJ/token)} &
\textbf{(mJ/token)} &
\textbf{(Wh)} \\
\midrule
EmbeddingGemma           & 0.308   & 0.961   & 1.153   & 0.000599 \\
MXBAI Embed Large        & 0.334   & 1.042   & 1.250   & 0.000696 \\
Qwen3 Embedding          & 0.600   & 1.872   & 2.246   & 0.001027 \\
Qwen3 (1.7B)             & 1.700   & 5.304   & 6.365   & 0.002641 \\
Granite 3.2 Vision       & 2.530   & 7.894   & 9.472   & 0.004923 \\
Qwen3 (8B)               & 8.000   & 24.960  & 29.952  & 0.011795 \\
Granite 3.3              & 8.170   & 25.490  & 30.588  & 0.012465 \\
Ministral 3 (14B)        & 14.000  & 43.680  & 52.416  & 0.021313 \\
DeepSeek-Coder V2        & 16.000  & 49.920  & 59.904  & 0.017425 \\
GPT-OSS (20B)            & 20.000  & 62.400  & 74.880  & 0.022281 \\
Qwen3 (32B)              & 32.000  & 99.840  & 119.808 & 0.052721 \\
Qwen2.5-Coder (32B)      & 32.000  & 99.840  & 119.808 & 0.052721 \\
Qwen3-VL (32B)           & 32.000  & 99.840  & 119.808 & 0.052721 \\
DeepSeek-R1              & 32.000  & 99.840  & 119.808 & 0.052721 \\
Llama 3.3 (70B)          & 70.000  & 218.400 & 262.080 & 0.170747 \\
GPT-OSS (120B)           & 120.000 & 374.400 & 449.280 & 0.155204 \\
\bottomrule
\end{tabular}
\end{table*}

Table~\ref{tab:model_inventory_token_energy} shows the expected linear dependence of token-level energy on model
size. Sub-billion-parameter embedding models have estimated token-level costs
below approximately 2.3 mJ/input token under the simplified estimator, whereas
70B- and 120B-parameter models require substantially higher per-token energy.
For example, the simplified estimate for LLaMA 3.3 70B is
218.4 mJ/output token and 262.1 mJ/input token under the short-prompt
assumption. The 120B model reaches 374.4 mJ/output token and
449.3 mJ/input token. These values are useful for comparative model selection because they expose the energy scaling of energy consumption with parameter count. The analysis isolates the cost of token processing and does not account for task performance. 

Models with identical parameter counts yield identical
token-level estimates under the parameter-scaled compute
model, since per-token FLOPs depend only on the number of
parameters. Architectural differences influence memory-related
energy and full request-level costs, but are not reflected in
these simplified token-level estimates. Finally, modality-specific models such as vision-language architectures may incur additional costs
that are not captured by the parameter-only formulation.

\subsection{Comparison with Measurement-Based Results}

Finally, we compare the analytical estimates against the measurement-based study
of Caravaca et al.~\cite{caravaca2025prompts}. The comparison uses the same
nominal workload:
\begin{equation}
T_{\mathrm{in}} = 500,
\qquad
T_{\mathrm{out}} = 500.
\end{equation}

\begin{table*}[t]
\centering
\caption{Analytical versus measured energy per prompt for a 500-token input and 500-token output workload.}
\label{tab:caravaca_comparison}

\small
\setlength{\tabcolsep}{12pt}
\renewcommand{\arraystretch}{1.1}

\begin{tabular}{lrrr}
\toprule
\textbf{Model size} &
\textbf{Analytical energy} &
\textbf{Measured energy} &
\textbf{Error} \\
\textbf{(B parameters)} &
\textbf{(Wh/request)} &
\textbf{(Wh/request)} &
\textbf{(\%)} \\
\midrule
8  & 0.011795 & 0.009270 & 27.23 \\
24 & 0.032427 & 0.026280 & 23.39 \\
70 & 0.170747 & 0.179230 & 4.73  \\
72 & 0.170747 & 0.233260 & 26.80 \\
\bottomrule
\end{tabular}
\end{table*}

Table~\ref{tab:caravaca_comparison} shows that the analytical estimator matches measurement-based values within approximately 5--27\% for the compared cases.
The lowest error is observed for the 70B model, where the analytical estimate differs from the measured value by 4.73\%. For the 8B, 24B, and 72B cases, the
relative error remains below 30\%. The remaining discrepancies are expected. The analytical model estimates GPU-level compute and memory movement under controlled assumptions, whereas
measurement-based studies include additional effects from inference engines,
batching policies, runtime scheduling, tensor-parallel execution, kernel fusion,
GPU utilization, and system-level overheads. Differences may also arise from
comparing models of similar size but different architecture. Therefore, the
comparison shows that the estimator captures the correct order of magnitude and scaling behavior, rather than as exact
per-deployment energy accounting.

Overall, the results indicate that the proposed analytical estimator provides a
transparent and reproducible approximation of LLM inference energy. It captures
the expected linear scaling with model size, the super-linear effect of
attention-related memory traffic in long generations, and the distinction
between input-token, output-token, and request-level energy.

\section{Limitations and Conclusion}

The proposed estimator is limited to accelerator-side operational energy. It
excludes CPU execution, host memory, networking, storage, power-supply losses,
cooling, and datacenter-level power usage effectiveness. Consequently, the
reported values are GPU-level operational energy estimates, not end-to-end service energy, datacenter energy, carbon emissions,
or lifecycle emissions. Nevertheless, accelerator-side energy is a major
controllable component in GPU-dense on-premise inference deployments: H100
SXM-class accelerators have a thermal design power of approximately 700 W, and
multi-GPU servers can therefore be dominated by accelerator power under high
utilization \cite{nvidia_h100_datasheet}. Facility-level energy remains larger
because of cooling and power-delivery overheads; for example, Uptime Institute
reports an average data-center PUE of approximately 1.56, and MLPerf Power
treats inference energy as a system-level measurement problem
\cite{uptime2024survey,mlperfpower2024,mlperfpowerdocs}. 

The estimator also abstracts away several architecture- and system-specific
effects. The parameter-scaled FLOP model does not fully capture feed-forward
expansion ratios, grouped-query attention, multi-query attention,
mixture-of-experts routing, or modality-specific processing in vision-language
models. The memory model approximates weight access, KV-cache writes, and
attention reads using calibrated factors $\gamma(N)$, $s_{\mathrm{attn}}(N)$,
and $\eta(N)$, which summarize parameter reuse, attention-access overhead, and
HBM inefficiency. These factors do not explicitly model the complete GPU memory
hierarchy, kernel scheduling, cache residency, batching behavior, or inference
engine implementation, and should be recalibrated for other hardware platforms,
serving engines, or batching regimes.

Despite these limitations, the methodology provides a reproducible framework
for comparing GPU-level inference energy across models and workloads when direct
instrumentation is unavailable. It also identifies actionable inference-stage
levers for reducing energy: smaller or task-specialized models reduce
parameter-scaled compute; shorter outputs reduce autoregressive decoding cost;
prompt compression and retrieval filtering reduce context length; quantized
weights and KV caches reduce memory traffic; and batching, prefix caching,
efficient attention kernels, speculative decoding, and model routing can improve
serving efficiency. The method is therefore best understood as a complementary
tool for green-coding analysis, comparative model selection, and design-time
evaluation, while precise deployment accounting still requires system-level
power measurement. 
Future work should integrate measured serving traces, extend the estimator to
quantized and mixture-of-experts models, and incorporate tensor-parallel
communication and batching effects.

\section*{Declaration on Generative AI}
During the preparation of this work, the authors used OpenAI ChatGPT for
language editing, restructuring, consistency checking, and drafting assistance.
After using this tool, the authors reviewed and edited the content as needed
and take full responsibility for the publication's content.
\section*{Acknowledgments} 
The authors gratefully acknowledge the ITEA GreenCode research project for fostering discussions on energy-aware AI systems and for funding parts of this work. The authors also thank their colleagues at TWT Science \& Innovation, Stefanos Papanikolaou and Michael Herrnberger, for their valuable discussions and insights.

\bibliography{bibliography}

\clearpage
\appendix

\section*{Supplementary Material}
\addcontentsline{toc}{section}{Supplementary Material}

\renewcommand{\thesection}{\Alph{section}}
\renewcommand{\thesubsection}{\thesection.\arabic{subsection}}
\renewcommand{\thesubsubsection}{\thesubsection.\arabic{subsubsection}}


\section{Derivation of the Compute Model}
\label{supp_sec:compute_derivation}

This section provides the architecture-aware compute terms used by the estimator.
The main paper reports the compact parameter-scaled form, while the full
attention correction terms are given here for reproducibility.

For dense decoder-only transformers, the dominant computation arises from matrix
multiplications in attention projections and feed-forward layers. When only the
model parameter count is available, the dense decode-phase FLOPs per generated
token, denoted $C_{\mathrm{dec/token}}$, are approximated as:
\begin{equation}
C_{\mathrm{dec/token}}
\approx
K N,
\label{supp_eq:decode_flops_per_token}
\end{equation}
where $N$ is the number of model parameters and $K$ is a transformer FLOP
constant. Following standard transformer FLOP accounting, we use:
\begin{equation}
K = 6,
\label{supp_eq:k_constant}
\end{equation} which corresponds to a commonly used approximation that dense
transformer inference requires on the order of six FLOPs per
parameter per token for the forward pass~\cite{kaplan2020scaling,hoffmann2022training}.

The dense compute cost of the decode phase is therefore:
\begin{equation}
C_{\mathrm{dec,dense}}
=
K N T_{\mathrm{out}}.
\label{supp_eq:decode_dense}
\end{equation}

Similarly, the dense compute cost of the prefill phase is:
\begin{equation}
C_{\mathrm{pre,dense}}
=
K N T_{\mathrm{in}}.
\label{supp_eq:prefill_dense}
\end{equation}

For long contexts, additional attention terms can be included. If
architecture-level parameters are available, the prefill attention correction is
approximated as:
\begin{equation}
C_{\mathrm{pre,attn}}
\approx
2 n_{\ell} d_{\mathrm{model}} T_{\mathrm{in}}^2,
\label{supp_eq:prefill_attention}
\end{equation}
where $n_{\ell}$ is the number of transformer layers and $d_{\mathrm{model}}$ is
the hidden dimension.

During decode, the token generated at step $t$ attends to a context of length
$T_{\mathrm{in}} + t - 1$. The attention-related decode cost is approximated as:
\begin{equation}
C_{\mathrm{dec,attn}}
\approx
4 n_{\ell} d_{\mathrm{model}}
\left(
T_{\mathrm{out}}T_{\mathrm{in}}
+
\frac{T_{\mathrm{out}}(T_{\mathrm{out}}-1)}{2}
\right).
\label{supp_eq:decode_attention}
\end{equation}

The total tensor-core compute estimate is:
\begin{equation}
C_{\mathrm{TC}}
=
C_{\mathrm{pre,dense}}
+
C_{\mathrm{dec,dense}}
+
C_{\mathrm{pre,attn}}
+
C_{\mathrm{dec,attn}}.
\label{supp_eq:total_compute}
\end{equation}

When detailed architectural parameters are unavailable, the attention correction
terms are omitted and the parameter-scaled approximation is used.

\section{Derivation of the Memory-Movement Model}
\label{supp_sec:memory_derivation}

This section provides the full memory-traffic decomposition used by the
architecture-aware estimator. The total HBM traffic is decomposed into parameter
access, KV-cache writes, and attention-related KV-cache reads:
\begin{equation}
\mathrm{Bits}_{\mathrm{total}}
=
\mathrm{Bits}_{\mathrm{params}}
+
\mathrm{Bits}_{\mathrm{KV}}
+
\mathrm{Bits}_{\mathrm{attn}}',
\label{supp_eq:bits_total}
\end{equation}
where $\mathrm{Bits}_{\mathrm{params}}$ denotes parameter-access traffic,
$\mathrm{Bits}_{\mathrm{KV}}$ denotes KV-cache write traffic, and
$\mathrm{Bits}_{\mathrm{attn}}'$ denotes scaled attention-related KV-cache read
traffic.

\subsection{Parameter-Access Traffic}

Model weights are stored in GPU memory and accessed during the forward passes
required for inference. A naive upper-bound formulation would assume that the
full set of model weights is transferred from HBM at every decoding step,
leading to memory traffic proportional to the number of generated tokens.
However, modern GPU implementations reduce this cost through on-chip caching,
kernel fusion, memory locality, and overlap between memory access and
computation.

To account for imperfect parameter reuse, we introduce a model-size-dependent
parameter-access factor $\gamma(N)$:
\begin{equation}
\mathrm{Bits}_{\mathrm{params}}
=
b_w N \gamma(N),
\label{supp_eq:bits_params}
\end{equation}
where $b_w$ is the number of bits per model weight and $N$ is the number of
model parameters.

The parameter-access factor $\gamma(N) \in (0,1]$ represents the effective
fraction of model parameters retrieved from HBM throughout the inference
request. We model its dependence on model size as:
\begin{equation}
\gamma(N)
=
\gamma_0
\left(
\frac{N}{N_0}
\right)^{\beta},
\label{supp_eq:gamma_scaling}
\end{equation}
where $N_0$ is a reference model size, $\gamma_0$ is the baseline reuse factor at $N_0$, and $\beta$ approximates the degradation of effective reuse as the model working set exceeds on-chip memory capacity.

\subsection{KV-Cache Write Traffic}

For each generated token, the model stores key and value vectors in each
transformer layer. The KV-cache write traffic is approximated as:
\begin{equation}
\mathrm{Bits}_{\mathrm{KV}}
=
2 b_{\mathrm{kv}} d_{\mathrm{model}} n_{\ell} T_{\mathrm{out}},
\label{supp_eq:bits_kv}
\end{equation}
where $b_{\mathrm{kv}}$ is the number of bits per KV-cache element,
$d_{\mathrm{model}}$ is the hidden dimension, $n_{\ell}$ is the number of
transformer layers, and $T_{\mathrm{out}}$ is the number of generated output
tokens. The factor of two accounts for storing both keys and values.

\subsection{Attention-Related KV-Cache Read Traffic}

During autoregressive decoding, each newly generated token attends over the
input prompt and the previously generated tokens through the KV cache. For an
input length $T_{\mathrm{in}}$ and an output sequence of length
$T_{\mathrm{out}}$, the baseline attention-related memory traffic is:
\begin{equation}
\mathrm{Bits}_{\mathrm{attn}}
=
2 b_{\mathrm{kv}} d_{\mathrm{model}} n_{\ell}
\left(
T_{\mathrm{out}}T_{\mathrm{in}}
+
\frac{
T_{\mathrm{out}}(T_{\mathrm{out}}-1)
}{2}
\right).
\label{supp_eq:bits_attn}
\end{equation}

This term includes a linear prompt-attention component and a quadratic
generated-token component. To account for non-ideal memory behavior, including irregular
access patterns, limited locality, cache contention, and increasing memory
pressure at larger model sizes, we apply an attention-specific scaling factor:
\begin{equation}
\mathrm{Bits}_{\mathrm{attn}}'
=
\mathrm{Bits}_{\mathrm{attn}}
\cdot
s_{\mathrm{attn}}(N),
\label{supp_eq:bits_attn_scaled}
\end{equation}

where $s_{\mathrm{attn}}(N) \geq 1$ is a dimensionless scaling factor.

The attention-related terms follow the quadratic dependence on
sequence length characteristic of standard self-attention, along
with the associated memory-access costs on GPU memory hierarchies
~\cite{dao2022flashattention}. 

The importance of KV-cache memory traffic in LLM serving has also been emphasized in prior system-level work, including PagedAttention and vLLM
~\cite{kwon2023efficient}.

\subsection{HBM-Dominated Memory Energy}

The memory-energy term is computed as:
\begin{equation}
E_{\mathrm{memory}}
=
\mathrm{Bits}_{\mathrm{total}}
\cdot
e_{\mathrm{HBM}}
\cdot
\eta(N),
\label{supp_eq:memory_energy_calibrated}
\end{equation}

where $e_{\mathrm{HBM}}$ is the energy per bit transferred from HBM and
$\eta(N) \geq 1$ is a global memory-inefficiency factor. The factor $\eta(N)$
approximates bandwidth saturation, memory-controller overhead, cache contention,
and pipeline stalls under high memory pressure.

The factors $\gamma(N)$, $s_{\mathrm{attn}}(N)$, and $\eta(N)$ are
empirical calibration terms fitted against measurement-based LLM
inference energy results reported by Caravaca et al.
~\cite{caravaca2025prompts}.

\section{Model Architecture Details}
\label{supp_sec:architeture_details}

In this section we present key architectural details of the models that have been used in this study. Table~\ref{supp_tab:model_presets} shows the parameter, layer, and hidden dimension ($d_{\mathrm{model}}$) of the model architectures.

\begin{table}[t]
\centering
\small
\caption{Overview of Model Architectural Configurations.}
\label{supp_tab:model_presets}

\begin{tabular}{lrrr}
\toprule
\textbf{Model} & \textbf{Params} & \textbf{Layers} & \textbf{$d_{model}$} \\
\midrule

Ministral 3 (14B) \cite{ministral3}             
& 14B & 40 & 5120 \\

EmbeddingGemma \cite{embeddinggemma}          
& 308M & 26 & 768 \\

MXBAI Embed Large \cite{mxbai}                
& 334M & 24 & 1024 \\

Qwen3 Embedding (0.6B) \cite{qwen3embed}      
& 0.6B & 28 & 1024 \\

Qwen3-VL (32B) \cite{qwen3vl}                 
& 32B & 64 & 5120 \\

Granite 3.3 (8B) \cite{granite33}             
& 8.17B & 40 & 4096 \\

Granite 3.2 Vision \cite{granite32}           
& 2.53B & 32 & 4096 \\

DeepSeek-Coder V2 (16B) \cite{deepseekcoder}  
& 16B & 27 & 2048 \\

DeepSeek-R1 (32B) \cite{deepseekr1}           
& 32B & 64 & 5120 \\

Qwen3 (8B) \cite{qwen38b}                     
& 8B & 36 & 4096 \\

Qwen3 (32B) \cite{qwen332b}                   
& 32B & 64 & 5120 \\

Qwen3 (1.7B) \cite{qwen31p7b}                 
& 1.7B & 28 & 2048 \\

Qwen2.5-Coder (32B) \cite{qwen25coder}        
& 32B & 64 & 5120 \\

Llama 3.3 (70B) \cite{llama33}                
& 70B & 80 & 8192 \\

GPT-OSS (120B) \cite{gptoss120}               
& 120B & 36 & 2880 \\

GPT-OSS (20B) \cite{gptoss20}                 
& 20B & 24 & 2880 \\

\bottomrule
\end{tabular}
\end{table}

\section{Reported Metrics}
\label{supp_sec:reported_metrics}

For each model--workload pair, the estimator reports energy at three levels of
granularity: phase-level energy, token-normalized energy, and aggregate
request-level energy. This separation is necessary because LLM inference is not
a homogeneous operation: prompt prefill and autoregressive decoding differ in
their compute structure, memory-access pattern, and dependence on sequence
length.

The phase-level quantities are the prefill energy,
$E_{\mathrm{pre}}$, and the decode energy, $E_{\mathrm{dec}}$. These terms
represent the estimated GPU energy required to process the input prompt and to
generate the output sequence, respectively. The total request energy is defined
as
\begin{equation}
E_{\mathrm{request}}
=
E_{\mathrm{pre}}
+
E_{\mathrm{dec}}.
\label{supp_eq:request_energy_metric}
\end{equation}

To compare requests with different prompt and generation lengths, we normalize
phase-level energy by the corresponding token counts. The input-token energy is
defined as
\begin{equation}
E_{\mathrm{in/token}}
=
\frac{E_{\mathrm{pre}}}{T_{\mathrm{in}}},
\label{supp_eq:input_token_metric}
\end{equation}
where $T_{\mathrm{in}}$ is the number of prompt tokens. Analogously, the
output-token energy is defined as
\begin{equation}
E_{\mathrm{out/token}}
=
\frac{E_{\mathrm{dec}}}{T_{\mathrm{out}}},
\label{supp_eq:output_token_metric}
\end{equation}
where $T_{\mathrm{out}}$ is the number of generated tokens.

We additionally report the average energy per processed token:
\begin{equation}
E_{\mathrm{avg/token}}
=
\frac{E_{\mathrm{request}}}
{T_{\mathrm{in}} + T_{\mathrm{out}}}.
\label{supp_eq:avg_token_metric}
\end{equation}

This aggregate metric mixes prefill
and decode costs and is therefore sensitive to the input/output token ratio.

Finally, to analyze the dominant sources of energy consumption, the request
energy is decomposed into compute and memory contributions:
\begin{equation}
E_{\mathrm{request}}
=
E_{\mathrm{compute}}
+
E_{\mathrm{memory}}.
\label{supp_eq:compute_memory_metric}
\end{equation}

The memory component is further attributed to parameter access, KV-cache writes,
and KV-cache reads. Reporting these components enables the evaluation to
distinguish compute-dominated regimes from memory-influenced or long-context
regimes.

\subsection{Interpretation of Simplified Token-Level Estimates}

The model-inventory table in the main paper reports simplified token-level
energy estimates based on a parameter-only, compute-dominated approximation:
\begin{equation}
\widehat{E}_{\mathrm{out/token}}
=
\alpha_{\mathrm{TC}} K N,
\qquad
\widehat{E}_{\mathrm{in/token}}
\approx
1.2\,\widehat{E}_{\mathrm{out/token}}.
\end{equation}

These values do not include the calibrated memory factors
$\gamma(N)$, $s_{\mathrm{attn}}(N)$, or $\eta(N)$ used in the
architecture-aware evaluation. They are
first-order comparative estimates across the model inventory, rather than
full compute-plus-memory request-level estimates. In contrast, the scaling and
decomposition figures in the main paper use the full calibrated
compute-plus-memory model.

\section{Implementation and Execution Assumptions}
\label{supp_sec:implementation}

This section describes the implementation details and system assumptions
underlying the analytical energy estimator.

\paragraph{Estimator implementation.}
The estimator is implemented as a Python-based analytical tool that
computes energy consumption from the model and the workload parameters.
Given the number of model parameters $N$, number of layers
$n_{\ell}$, hidden dimension $d_{\mathrm{model}}$, and token
counts, the tool calculates FLOPs and memory traffic using the
analytical formulas described in the main text.

The implementation does not execute neural networks or perform
runtime profiling. Instead, it deterministically estimates energy
based on compute and memory abstractions. No direct GPU telemetry, wall-plug power instrumentation, or runtime energy
measurement APIs (e.g., NVML or nvidia-smi) are used during estimation.

\paragraph{Hardware assumptions.}
The main evaluation uses H100-class coefficients, while additional accelerator
coefficients are reported here for completeness. Energy conversion coefficients
for tensor-core operations ($\alpha_{\mathrm{TC}}$) and HBM traffic
($e_{\mathrm{HBM}}$) are taken from measurement-based prior
work~\cite{antepara2025benchmark}.

\begin{table}[th]
\centering
\small
\caption{
Accelerator-level energy coefficients for FP16/BF16 tensor-core inference.
}
\label{supp_tab:hardware_coefficients}
\begin{tabular}{lcc}
\toprule
\textbf{Hardware} &
\textbf{$\alpha_{\mathrm{TC}}$ (pJ/FLOP)} &
\textbf{$e_{\mathrm{HBM}}$ (pJ/bit)} \\
\midrule
H100 / GH200 & 0.52 & 11.68 \\
A100         & 0.70 & 13.11 \\
\bottomrule
\end{tabular}
\end{table}

\paragraph{Software stack assumptions.}
The estimator assumes inference execution on optimized tensor-core GPU kernels
using FP16/BF16 arithmetic and HBM-resident model weights. It is intended to
approximate inference behavior of optimized GPU-based implementations built on
CUDA and high-performance libraries such as cuBLAS and cuDNN, as well as modern
LLM inference frameworks (e.g., TensorRT-LLM, Megatron-LM, and vLLM).

The model does not explicitly simulate kernel-level execution or
software-specific optimizations such as kernel fusion, scheduling,
or memory tiling. Instead, these system-level effects are treated
implicitly and are approximated through the empirical scaling
factors $\gamma(N)$, $s_{\mathrm{attn}}(N)$, and $\eta(N)$, which
collectively capture deviations from ideal compute and memory
behavior observed in optimized inference systems.

\paragraph{Calibration reference.}
Model parameters are calibrated using reported energy measurements
for optimized LLM inference from prior work~\cite{caravaca2025prompts}.
These measurements serve as reference values for matching the
magnitude and scaling behavior of energy consumption.

\paragraph{Limitations.}
The estimator does not model GPU execution at the kernel or
instruction level. In particular, it does not simulate thread-level
parallelism, CUDA scheduling, or detailed memory hierarchy behavior.
Instead, such effects are approximated through calibrated scaling
factors.

As a result, the model is an analytical approximation rather than a cycle-accurate simulation or direct hardware measurement.

\end{document}